\pdfoutput=1
\def\year{2019}\relax
\documentclass[letterpaper]{article} %DO NOT CHANGE THIS
\usepackage{aaai19}  %Required
\usepackage{times}  %Required
\usepackage{helvet}  %Required
\usepackage{courier}  %Required
\usepackage{url}  %Required
\usepackage{graphicx}  %Required
\usepackage{subcaption}
\usepackage{amsmath}
\newcommand{\citet}[1]
{\citeauthor{#1}˜\shortcite{#1}}
\newcommand{\citep}{\cite}

\begin{document}
% The file aaai.sty is the style file for AAAI Press 
% proceedings, working notes, and technical reports.
%
\title{DA-LSTM: A Long Short-Term Memory with Depth Adaptive to \\ Non-uniform Information Flow in Sequential Data}
\author{
	Yifeng Zhang, Ka-Ho Chow, S.-H. Gary Chan \\
	Department of Computer Science and Engineering\\
	Hong Kong University of Science and Technology\\
	yzhangch@cse.ust.hk, khchowad@cse.ust.hk, gchan@cse.ust.hk
}
\maketitle
\begin{abstract}
Much sequential data exhibits highly non-uniform information distribution.  This cannot be correctly modeled by traditional Long Short-Term Memory (LSTM). To address that, recent works have extended LSTM by adding more activations between adjacent inputs.
%Such deeper structures equip LSTMs with stronger capabilities to learn data statistics, including long-term dependencies and attention, thereby achieving state-of-the-art performance.
However, the approaches often use a fixed depth, which is at the step of the most information content. This one-size-fits-all worst-case approach is not satisfactory, because
when little information is distributed to some steps, shallow structures can achieve faster convergence and consume less computation resource.

In this paper, we develop a Depth-Adaptive Long Short-Term Memory (DA-LSTM) architecture, which can dynamically adjust the structure depending on information distribution without prior knowledge.
%adapt model depth to non-uniform information flow 
%Therefore, devising a flexible model that , can save computation load without degrading performance.
Experimental results on real-world datasets show that DA-LSTM costs much less computation resource and substantially reduce convergence time by $41.78\%$ and $46.01 \%$, compared with Stacked LSTM and Deep Transition LSTM, respectively. 
\end{abstract}

\section{Introduction}
\begin{figure}
\centering
\includegraphics*[]{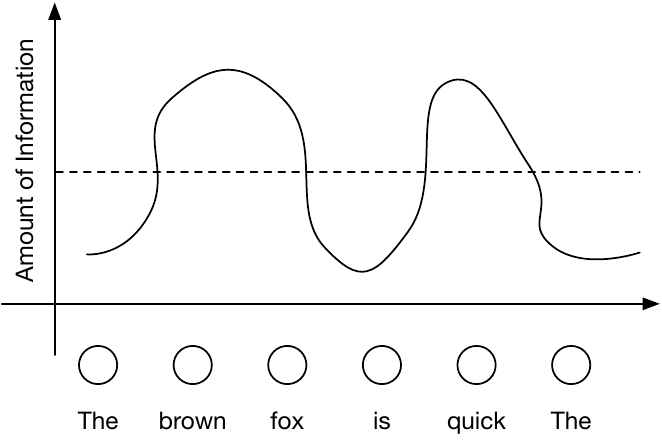}
\caption{Non-uniform information flow in a text sequence.} \label{fig:nonuniform}
\end{figure}

There are many examples for sequential data with non-uniform information flow. One is text data, where nouns and adjectives can be more crucial than determiners to modify sentence context.  We show such a case in Figure~\ref{fig:nonuniform}. Words "brown", "fox" and "quick" contain larger amount of information than "the" and "is".  Yet another example is video sequence, where the difference between adjacent video frames varies. In other words, some video frames can change drastically to convey quantities of information while others change slightly or even freezing. In sensor systems, sequential data may also be non-uniform. As different sensors have diverse sampling rates, accompanying with occasional package loss in wireless connection, the resulting frequency of the signal sequence is highly variable, leading to a non-uniform information distribution.

To process sequential data,
Recurrent Neural Networks (RNNs) \cite{rumelhart1986learning} have
emerged as a popular approach.  It has been applied
for speech recognition \cite{graves2013speech},
language modeling \cite{sundermeyer2012lstm}, learning word
embeddings \cite{goldberg2014word2vec}, location
prediction \cite{kong2018hst}, electronic health record
analysis \cite{jin2018predicting}, etc. Among all kinds of RNNs and
their extensions, Long Short-Term Memory
(LSTM) \cite{hochreiter1997long} is immensely important to learn
long-term dependencies.  This is because of its gated structure, which
alleviates the problem of vanishing and exploding gradients.

% what is depth and how most works do
Although traditional LSTM has produced promising results,
it performs the best for
data sequences where information is uniformly distributed between steps.
For sequences with non-uniform information flow between steps,
its performance is hardly satisfactory.
The work in \cite{pascanu2013construct} has investigated the case and attributed the problem to the lack of depth between steps in its design. In other words, for high information flow, the transition function from one hidden state to the next is too shallow in LSTM to the point that it may be regarded as a single linear transformation with activation, which cannot capture the latent structure in sequences.

To tackle the above, many recent works devise deeper structures such as adding intermediate layers or stacking LSTMs.  An approach is Fast-Slow LSTM (FS-LSTM) \cite{mujika2017fast}, which forms a two-layer hierarchical structure. The approach uses a
fixed ``worst-case'' depth according to the maximum information flow in a step in the data sequence. This is inefficient.
%The major drawback of such approach is efficiency.
For example, in speech recognition, little operations are needed for modeling silent parts, and hence applying a deep LSTM structure can waste computation resources. Besides, a deep structure can possibly lead to vanishing and exploding gradients because error signals have to traverse a long path.
Therefore, models are required to handle such sequences with adaptive depth in each step.

% not always good, an example of sequences

Phased LSTM has been proposed to address the above \cite{neil2016phased}, which extends LSTM with a time gate whose openness is controlled by an independent rhythmic oscillation. Based on the temporal information, information flow is normalized, thereby improving the performance. A similar implementation is Heterogeneous Event LSTM (HE-LSTM) \cite{liu2018learning}, which replaces time gate with an event gate. The change enables learning on the hierarchical structure of input features and produces a better result in asynchronous sequential data. However,
%there are two shortcomings to the above two approaches. Firstly,
both methods require prior knowledge on the latent structure of sequences, which may not be known in reality.
%can be inaccessible in real-work tasks. Secondly,
Furthermore, for sequences without timestamp features, normalization cannot be conducted.

In this paper, we propose Depth-Adaptive Long Short-Term Memory (DA-LSTM), a hierarchical architecture that can adapt model depth to non-uniform information flow in sequential data. Different from Phased LSTM and HE-LSTM, DA-LSTM neither requires any prior knowledge nor timestamps of sequences. DA-LSTM consists of two layers. In the bottom layer, multiple cells are sequentially connected to increase the capability of modeling multiscale transition, while in the top layer, a short path is constructed by linking the  head and tail cell in the bottom layer, alleviating gradients exploding and vanishing problem. Additionally, we add a portion gate into the LSTM cell, which can dynamically determine the size of hidden units to be updated. Therefore, DA-LSTM can provide sufficient depth for information-dense steps, while saving computation load at information-sparse inputs.

Using DA-LSTM, we conduct experiments on real-world classification tasks. Experimental results on the dataset show that DA-LSTM can dynamically adjust model depth, and hence achieves faster training speed while preserving, or even improving baseline performances without prior knowledge of sequences' latent structure. Therefore, DA-LSTM can be universally applied to process sequences with non-uniform information distribution.

The remaining parts are organized as follows. Firstly, related works are presented and compared with DA-LSTM.  Next, we exhibit the hierarchical structure and cell structure of our model. We then propose experimental results and illustrations, followed by a comprehensive conclusion. 

\section{Related Works}
In this section, we review works that are relevant to DA-LSTM. Firstly, we focus on methods that have a deep structure to model complicated transition, which contains a considerable amount of information. Next, several approaches that can flexibly adjust models are compared. Finally, we discuss how our model differs from these two types of methods.

Standard LSTM can sometimes generate suboptimal results. Works have investigated the problem and shown that the degradation in performance can be attributed to the shallow hidden-to-hidden depth \cite{pascanu2013construct}. In other words, because only a few linear and nonlinear transformations are performed between hidden states at adjacent time steps, standard LSTM is not capable of modeling complex features, such as multiscale temporal structure or localized attention. Therefore, many works have increased the model depth to boost performance. Deep Transition LSTM \cite{pascanu2013construct} is one typical approach, which inserts several intermediate layers between hidden layers. However, adding too many intermediate layers introduces a potential problem. As the number of nonlinear steps increases, gradients should traverse a longer path during backpropagation, contributing to the problem of vanishing and exploding gradients. Recurrent Highway Networks (RHN) \cite{zilly2016recurrent} address the issue by utilizing the highway layer \cite{srivastava2015highway}, which achieves state-of-the-art BPC measure on Penn Treebank and Hutter Prize Wikipedia datasets. Stacking recurrent hidden layers is another primary approach. Models of this concept have been applied in sequence segmentation and recognition tasks \cite{graves2013speech}, achieving baseline results on the TIMIT phoneme recognition benchmark. More recent works introduce the hierarchical design, such as Clockwork RNN (CW-RNN) \cite{pmlr-v32-koutnik14}, which divides hidden states into several modules with different clock-rate. The model follows an asynchronous updating rule and is capable of capturing multiscale latent structure over sequences.

However, the better performance of models with an increased depth is at the expense of higher computation load and harder training process. Since the sequence can have a highly variable amount of information, models are not necessary to maintain equal depth over the whole sequence. Models that have a mechanism to adjust depth according to sequences' latent structure can significantly save computation resources. One type of models use timestamp as an indicator to handle non-uniform sequences, such as Phased LSTM\cite{neil2016phased} and Time-Aware LSTM\cite{baytas2017patient}. The former model follows a periodic oscillation function to control the hidden state transition while the latter one discounts transition with an arbitrary function. One limitation of these models is that they both require some prior knowledge about sequences' temporal structure, which cannot be obtained in many scenarios. Another implementation is adaptive computing time (ACT) \cite{graves2016adaptive}, which processes a varied number of intermediate updates between steps. ACT utilizes a halting gate to assign each update with a probability. When the sum of prior halting probabilities exceeds $1-\epsilon$, ACT stops updating at the current step and enter the next step, resetting cumulated halting probability to zero. Implementation of a similar concept is Hierarchical Multiscale Recurrent Neural Networks (HM-LSTM) \cite{chung2017hierarchical}. This work develops a multi-layer LSTM, which has both bottom-up and top-down connections. Each layer in HM-LSTM has three possible operations, namely COPY, UPDATE and FLUSH. A boundary state decides the choice of operations, whose design is borrowed by Skip-RNN \cite{campos2018skip}. Despite their high performance, both methods need to be trained by REINFORCE \cite{williams1992simple} and require an extra reward signal. Defining these rewards is demanding and task-specific. Variable Computation Unit (VCU) \cite{jernite2017variable} provides an alternative view for adjusting depth. Instead of skipping the whole cell as Skip-RNN, VCU performs partial updates, leaving higher dimensions in hidden units intact. 

\begin{figure*}
\centering
\includegraphics*[]{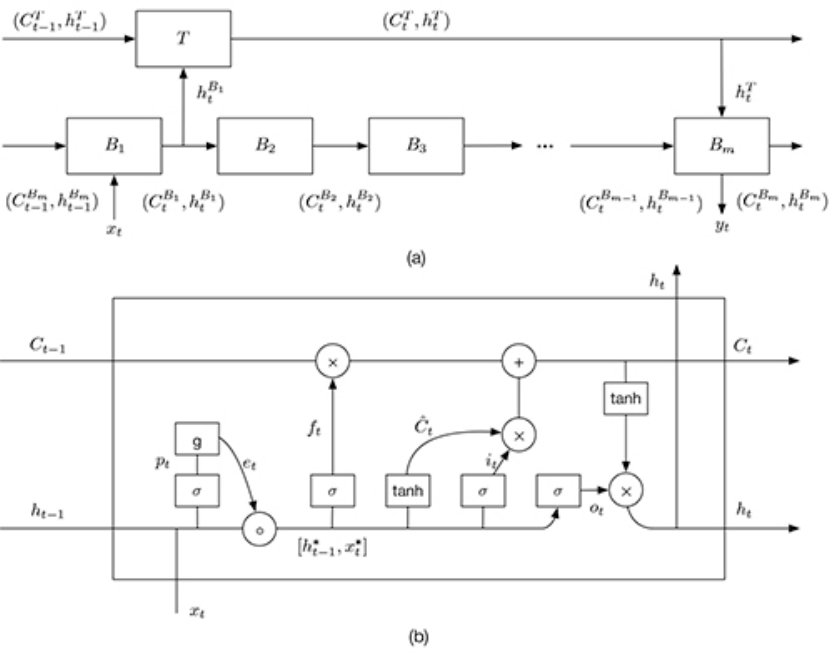}
\caption{Proposed DA-LSTM architecture (a) Hierarchical structure (b) Modified LSTM cell.} \label{fig:methodstructure}
\end{figure*}

Our DA-LSTM architecture combines the advantages of both deep structure and flexible depth adjustment mechanism. Firstly, by forming a two-layer hierarchical structure, DA-LSTM is equipped with sufficient depth to model multiscale complex transitions. Secondly, since a portion gate is added to the standard LSTM cell, our model is flexible enough to assign an appropriate amount of operations for hidden state updates, reducing computation load and speeding up training.

\section{DA-LSTM Architecture}
In this section, we first introduce the hierarchical structure of DA-LSTM to obtain an overview. Then, the modified LSTM cell is displayed in details to illustrate how the portion gate helps to save computation resources.
\subsection{DA-LSTM Hierarchical Framework}
Transitions in some parts of the sequential data can be complex, such as periods when  changes between steps are drastic, containing a large amount of information, and thereby require models to be capable of capturing various latent features. Since a single standard LSTM cell has a shallow structure between hidden states, which only consists of several linear transformations and activations, learning statistics from sequences is challenging. To solve the problem caused by insufficient depth, we propose a hierarchical structure (see figure \ref{fig:methodstructure}(a)), which comprises two stacked layers.

In the bottom layer, $m$ cells $B_1,B_2,B_3,...,B_m$ are connected sequentially. Each cell can represent a transition function $f_t^{B_i}(h,x)$ at time step t, and therefore this layer can model complex transition by processing multiple updates $f_t^{B_m}(f_t^{B_{m-1}}(...f_t^{B_{1}}(h,x)...))$ within one step. Here $x$ can be input from sequences or the output from cells in different layers. If there is no additional input, $x$ can be omitted by setting the value to zero. The tuple of memory and hidden states is represented by $h$. Only the first cell $B_1$ receives an input and the last cell $B_m$ emits an output. The top layer consists of one customized cell $T$, which is updated once at every step. Cell $T$ receives the signal from the first cell $B_1$ and sends feedback to the last cell $B_m$ in the lower layer. Equations about how information transmits in DA-LSTM architecture is presented in the following:
\begin{align}
(C_t^{B_1}, h_t^{B_1}) &= f^{B_1}(C_{t-1}^{B_m},h_{t-1}^{B_m},x_t) \\
(C_t^T, h_t^T) &= f^{T}(C_{t-1}^T, h_{t-1}^T, h_t^{B_1}) \\
(C_t^T, h_t^T) &= f^{T}(C_{t-1}^T, h_{t-1}^T, h_t^{B_1}) \\
(C_t^{B_i}, h_t^{B_i}) &= f^{B_i}(C_{t}^{B_{i-1}}, h_{t}^{B_{i-1}})~~i\in [2,m-1] \\
(C_{t}^{B_m}, h_t^{B_m}) &= f^{B_m}(C_{t}^{B_{m-1}}, h_{t}^{B_{m-1}}, h_{t}^T) \\
y_t &= o(h_t^{B_m}).
\end{align}
Note that the function $o$ in (5) can be any transformation, converting hidden states $h_t^{B_m}$ to a predicted output. 

Compared with Deep Transition LSTM, DA-LSTM maintains a two-layer hierarchical structure with $1$ and $m$ cells, respectively. While the bottom layer is similar to Deep Transition LSTM, the top layer is updated with a different frequency, enabling the learning of multiscale latent features. Besides, links between layers can be regarded as a shortcut connection \cite{raiko2012deep}, which alleviates the problem of vanishing and exploding gradients. Therefore, error signals can be backpropagated for longer steps, and the learning of long-term dependencies is promoted. DA-LSTM also differs from stacked LSTM. Instead of connecting layers with only bottom-up links, DA-LSTM also contains top-down links. Therefore, The hidden state transitions in DA-LSTM can be regarded as multilayer perceptron (MLP), which is a universal approximator \cite{hornik1989multilayer} and capable of representing larger families of functions.

\begin{figure}
\centering
\includegraphics*[]{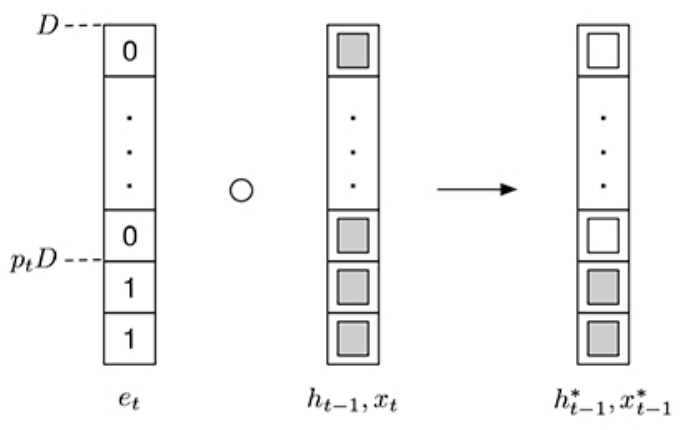}
\caption{Illustration of how hidden states are truncated.} \label{fig:partialupdates}
\end{figure}
\subsection{Modified LSTM Cell Structure}
Long Short-Term Memory (LSTM)\cite{hochreiter1997long} cell is a pervasive unit in building recurrent neural networks. We start by defining update rules with equations for LSTM without peephole connection \cite{gers2000recurrent}:
\begin{align}
f_t &= \sigma(W_fh_{t-1}+U_fx_{t}+b_f) \\
i_t &= \sigma(W_ih_{t-1}+U_ix_{t}+b_i) \\
o_t &= \sigma(W_oh_{t-1}+U_ox_{t}+b_o) \\
\hat{C_t} &= tanh(W_ch_{t-1}+U_cx_{t}+b_C) \\
C_t &= f_t*C_{t-1}+i_t*\hat{C}_t \\
h_t &= o_t*tanh(C_t).
\end{align}
Compared with RNN cells, LSTM utilizes the gated structure $f_t$, $i_t$ and $o_t$, namely forget gate, input gate and output gate at time step $t$. $C_t$ is the new memory candidate, which is computed by a linear transformation of previous hidden state $h_{t-1}$ and new input $x_t$ concatenated with activation function $tanh$. $W_f$, $W_i$, $W_o$, $W_C$, $U_f$, $U_i$, $U_o$, $U_C$ and $b_f$, $b_i$, $b_o$, $b_C$ are parameters in corresponding gates. The forget gate and input gate, which choose sigmoid function as nonlinearity, output scalar values. The values are then multiplied with previous memory $C_{t-1}$ and new memory candidate $\hat{C}_t$, respectively, to produce a new memory $C_t$. Finally, output $h_t$ is obtained by the element-wise multiplication between $tanh(C_t)$ and the value of output gate $o_t$.

The modified cell in DA-LSTM extends LSTM with a portion gate $p_t$, which defines a faction number within interval $(0,1)$. As is shown in the following, $W_p$, $U_p$ and $b_p$ are parameters for portion gate with sigmoid activation.
\begin{align}
p_t &= \sigma(W_ph_{t-1}+U_px_t+b_p).
\end{align}
Figure \ref{fig:methodstructure}(b) shows the whole structure of modified LSTM cell. Assume previous hidden state $h_{t-1}$ has a dimension of $D$, DA-LSTM will use the first $\lceil p_tD \rceil$ dimensions of hidden units and $x_t$. Note that here $x_t$ can either be the input for the cell in the top layer or a new input in the bottom layer. In the former case, $x_t$'s dimension is $D$ because the input for cell $T$ is exactly the hidden state output $h_t^{B_1}$ of cell $B_1$, which has the same dimension as $h_{t-1}^{B_m}$. In the latter case, the dimension of $x_t$ can be regarded as $D$ for simplification, since an extra preprocessing layer can be inserted to transform the dimension of raw input to $D$. Therefore, $x_t$ is assumed to have the same dimension as $h_{t-1}$. 

After obtaining $p_t$, the modified LSTM cell performs partial update on dimension $[1, \lceil p_tD \rceil]$. As can be seen from figure \ref{fig:partialupdates}, dimensions over $\lceil p_tD \rceil$ are truncated, generating $h_{t-1}^*$ and $x_t^*$, hence only $O(p_t^2D^2)$ multiplications are needed to compute the rest gates. Since the multiplication account for the majority of computation load, time and resources are expected to be saved. 

In practice, hard selection of the updating dimension size is not favored because training model with non-differentiable function is non-trivial. Here we apply a soft mask \cite{jernite2017variable}, which approximates the hard choice with a threshold function. We use $g$ in figure \ref{fig:methodstructure}(b) to denote the continuous function that generates a soft mask vector. The threshold function is defined as (element-wise multiplication is denoted by $\circ$):
\begin{align}
Thres_{\epsilon}(x) &= \begin{cases}
0,\qquad x<\epsilon \\
1,\qquad x>1-\epsilon \\
x,\qquad otherwise.
\end{cases}
\end{align}
Given a sharpness parameter $\lambda$, the soft mask vector $e_t$, truncated hidden state $h_{t-1}^*$ and truncated input $x_t^*$ are defined as:
\begin{align}
\forall i\in {1,2,...,D}, (e_t)_i &= Thres_\epsilon(\sigma(\lambda(p_tD-i))) \\
h_{t-1}^* &= h_{t-1} \circ e_t \\
x_{t}^* &= x_t \circ e_t.
\end{align}
The element-wise multiplication of soft mask $e_t$ and $[h_{t-1},x_t]$ generates $[h_{t-1}^*, x_t^*]$, whose first $(p_tD+\tau)$ dimensions are unchanged, with dimensions over $((1-p_t)D-\tau)$ being zero. Here $\tau$ is the offset caused by the fraction part of $p_tD$. If the sharpness parameter $\lambda$ increases, $\tau$ will decrease to zero because the value of $\sigma(\lambda(p_tD-i)$ is less likely to fall into the range between $\epsilon$ and $1-\epsilon$.

\section{Experiments and Evaluation}
In this section, experimental results on real sensor data are performed to prove the efficiency and effectiveness of our proposed DA-LSTM architecture. 
\subsection{Datasets and Experimental Setup}
The DA-LSTM architecture is tested on PAMAP2 Physical Activity Monitoring dataset \cite{reiss2012introducing} \cite{reiss2012creating}, which can be obtained from UCI Machine Learning Repository. PAMAP2 is collected by putting wearable compounded sensors over a group of people aged between $24$ and $30$ for $10$ hours, entirely providing $3,850,505$ instances. The specification of sensors can be seen in Table \ref{table:sensorspec}. 

Since thermometer, 3D-accelerator, 3D-gyroscope, and 3D-magnetometer are embedded into a compound sensor called IMU sensor, they share the same sampling rates. However, as sensors communicate with wireless connection, the real frequency of IMU sensors placed at different body positions varies and package loss frequently occurs, generating quantities of NaN tags in resulting integrated records. We firstly preprocess sequential data by removing rows of signal records with NaN tag. All the sensors provide $52$ features for physical activity recognition.

There are $25$ activity types, including sitting, walking, running, which are more recognizable, and house cleaning, folding laundry, ironing, which are less distinguishable. Besides, states between these activities, namely transient states, is labeled as $0$, where sensor readings are noisy and irrelevant to predictions. Therefore, these transient periods can be regarded as information-sparse parts, and we define a transient ratio $r$ to control the ratio of transient states in our extracted sequences. If $r$ is too large, inputs will mostly consist of transient states and information will be uniformly distributed. Similarly, when $r$ is too small, inputs will contain quantities of non-transient states, whose information distribution is still uniform. The only difference is the total amount of information distributed. In other words, sequences with small $r$ contain a larger number of complicated transitions while sequences with large $r$ contain a smaller number of complicated ones. If $r$ is near $0.5$, however, the amount of information distributed at different steps are highly variable.

We follow these steps to extract sequences from 10-hour sensor records. Firstly, records are split into two parts, one with nothing but transient states and the other with rest records. Next, we follow a transient ratio $r$ to sample data points from different parts. Finally, records for transient states and non-transient states from the same test objects are concatenated and sorted in chronological order, with a sequence length parameter $n$ defining number of steps in each sequence. Unless otherwise stated, we use the following baseline parameters: $\{r=0.5,\ n=200\}$. Besides, The proportions of the training set, the cross-validation set, and the testing set are $80\%$, $10\%$ and $10\%$, respectively. Several single cell and hierarchical structures are tested on these extracted sequential data, and all the experiments are conducted with tensorflow framework.
\begin{table}
\centering
\caption{Sensor specifications.} \label{table:sensorspec}
\begin{tabular}{|c|c|c|c|}
\hline
sensor & number & frequency \\
\hline
thermometer & 3 & 100Hz \\
\hline
3D-accerator & 6 & 100Hz \\
\hline
3D-gyroscope & 3 & 100Hz \\
\hline
3D-magnetometer & 3 & 100Hz \\
\hline
Heart Rate Monitor & 1 & ~9Hz \\
\hline
\end{tabular}
\end{table}
\subsection{Comparison Schemes and Metrics}
DA-LSTM architecture is compared with the following methods to prove its effectiveness and efficiency. 
\begin{itemize}
\item Phased LSTM: Phased LSTM \cite{neil2016phased} as described in related work.
\item Stacked LSTM: LSTM cells are stacked on the top of each other by treating the output of the bottom layer as input \cite{malhotra2015long}. As for this structure, there is no link conveying the information from the upper cell to lower cell.
\item Deep Transition LSTM(DT-LSTM): A model that sequentially connects several cells between steps \cite{raiko2012deep}.
\item Clockwork RNN: Clockwork RNN \cite{pmlr-v32-koutnik14} as described in related work.
\end{itemize}
We choose the default parameter settings for Phased LSTM and Clockwork RNN from corresponding papers. Other common hyperparameters are listed in table \ref{table:hyperparameter}. Note that modules in Clockwork RNN contain different number of cells and we use addition to denote the structure.
\begin{table}
\centering
\caption{Hyperparameters of comparing methods.} \label{table:hyperparameter}
\begin{tabular}{|c|c|c|}
\hline
method & \# of hidden units & \# of cells \\
%\hline
%LSTM  & 120 & 1 \\
%\hline
%Modified LSTM & 120 & 1 \\
\hline
Phased LSTM & 120 & 1 \\
\hline
Stacked LSTM & 40 & 3  \\
\hline
Deep Transition LSTM & 40 & 3 \\
\hline
Clockwork RNN & 40 & 1+2 \\
\hline
DA-LSTM & 40 & 3 \\
\hline
\end{tabular}
\end{table}
Since all the methods are tested on classification tasks, it is common to use cross entropy loss as a quantitative metric. Besides, convergence time is recorded to indicate the computation load.

\begin{table*}
\centering
\caption{Performance of activity classification task with ${r=0.5}$.} \label{table:result}
\begin{tabular}{|c||c|c|c|}
\hline
Methods & Cross Entropy at Epoch 1 & Cross Entropy & Convergence Time (s) \\
\hline
Phased LSTM &  1.4743 & 0.9336 &  1316.36 \\
\hline
Stacked LSTM & 1.3713  & 0.4736  & 8287.75 \\
\hline
Deep Transition LSTM & 1.3609 & 0.7932 &  8935.57 \\
\hline
Clockwork RNN & 1.3224 & 0.4807 &  1604.24 \\
\hline
DA-LSTM & \textbf{1.3671} & \textbf{0.5284} & \textbf{4824.20} \\
\hline
\end{tabular}
\end{table*}

\subsection{Illustrative Results}
Table \ref{table:result} lists cross entropy loss in the test set and convergence time with baseline parameters. Several conclusions can be drawn from the table. 

Firstly, it can be seen that Clockwork LSTM and DA-LSTM achieve lower cross entropy loss than Deep Transition LSTM and Phased LSTM. The performance of Deep Transition LSTM is degraded because gradients should traverse a longer path in backpropagation, which hampers the learning of long-term dependencies. As for Phased LSTM, suboptimal results can be attributed to the incompatibility between the real latent structure in sequences and Phased LSTM's temporal assumption. Therefore, we can conclude that both longer backpropagation paths and false assumption about sequences' latent structure are detrimental to performance.

% DA-LSTM save computation significantly
Secondly, comparing Phased LSTM with Stacked LSTM, Deep Transition LSTM and DA-LSTM, it can be seen that the convergence time for the single cell structure is significantly smaller than hierarchical structures. Instead of forming multilayer cells between steps, Phased LSTM maintains a simpler structure, performing much less computation. However, even though hierarchical structures require higher computation load, DA-LSTM manages to save the cost. Compared with Stacked LSTM and Deep Transition LSTM, DA-LSTM reduces convergence time significantly by introducing a portion gate, which can dynamically adjust model depth  and omit quantities of operations according to the latent structure of sequences. Therefore, a conclusion that DA-LSTM is capable of saving computation resources can be drawn. Here the time for Clockwork RNN cannot be fairly compared because its basic unit is not LSTM cell, which reduces lots of overheads and operations. 

\begin{figure}
\centering
\includegraphics*[]{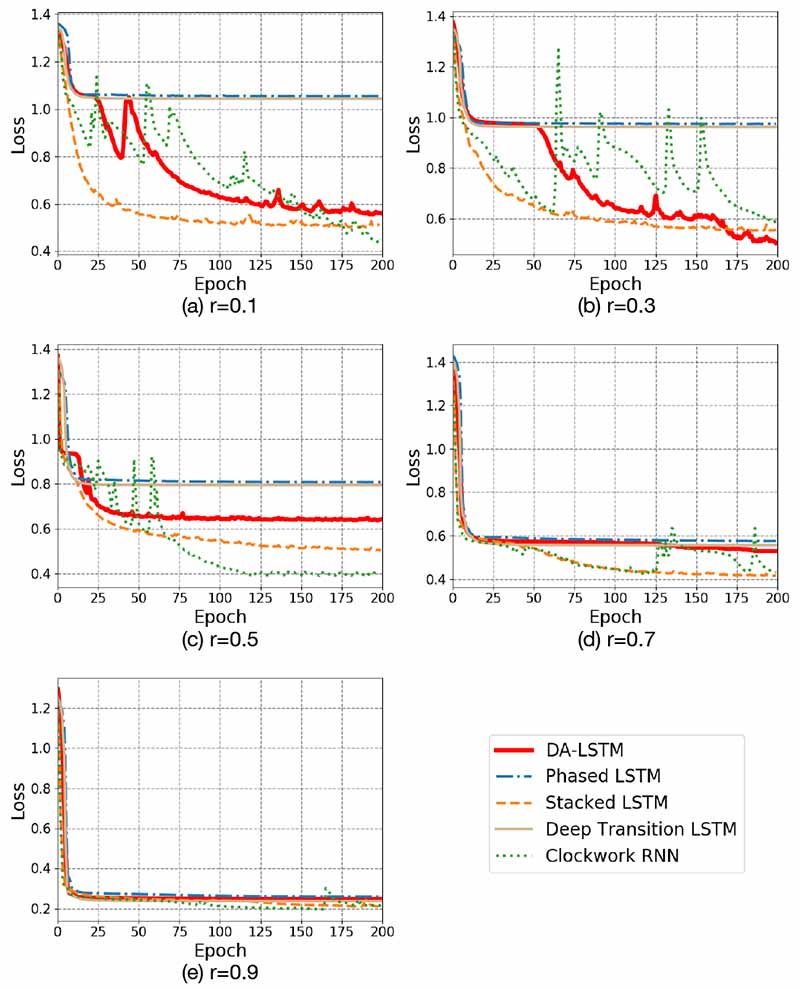}
\caption{Learning curves over different transient ratio $r$.} \label{fig:learningcurve}
\end{figure}

% compare with phased LSTM / CW-RNN
Thirdly, it can be seen that both DA-LSTM and Clockwork RNN achieve state-of-the-art performance. Different from Clockwork RNN, whose module settings implicitly define some temporal structure information of inputs, DA-LSTM does not require any prior knowledge about inputs' latent structure. Even though the temporal assumption formed by Clockwork RNN is more general than Phased LSTM, achieving better performance, when we modify the transient ratio $r$, results of Clockwork RNN become quite unstable. Details will be provided in the following parts.

% r, performance (DA-LSTM is more robust) % CW-RNN and 
To further display DA-LSTM's capabilities of dealing with non-uniform information flow in sequential data, we run experiments on sequences with different transient ratio $r$. Figure \ref{fig:learningcurve} displays corresponding learning curves.
From these graphs, when we change the transient ratio $r$, the resulting cross entropy of Clockwork RNN fluctuates rapidly. It can be seen that Clockwork RNN generates suboptimal results when $r=0.3$, which is possibly due to the incompatibility between the model's assumption and sequences' latent structure. DA-LSTM, however, always maintains a stable performance.

% Deep transition LSTM and Phased LSTM always perform badly
Besides, we find that the performance of both Deep Transition LSTM and Phased LSTM is inferior. As for Phased LSTM, because we use default parameter settings described in the corresponding paper, the assumption for the latent temporal structure of inputs may not be congenital with our datasets, hence generating suboptimal results. The bad performance of Deep Transition LSTM can be attributed to a different reason. Although adding intermediate layers between cells increase the depth of deep transition LSTM, which is supposed to provide better performance, a longer backpropagation path intensifies the problem of vanishing and exploding gradients. Therefore, Deep Transition LSTM is incapable of capturing long-term dependencies. DA-LSTM alleviates the problem by applying links to connect the start and end of intermediate cells, allowing gradients to traverse a short path and thereby promoting the learning of long-term dependencies. One exception occurs when $r=0.9$. Because the latent structure of sequential inputs is quite simple at that time, all the methods achieve high accuracy.

Based on the above experimental results, we can conclude that DA-LSTM architecture can significantly save computation cost while preserving or sometimes even improving the state-of-the-art performance.

\subsection{Depth-Adaptive Study}
In this section, to understand the how DA-LSTM adapts depth, we run experiments with different transient ratio $r$ and cell number $m$. The average portion gate value $p$ after convergence is recorded for comparison.

\begin{figure}
\begin{subfigure}{0.2\textwidth}
\centering
\includegraphics[width=\linewidth]{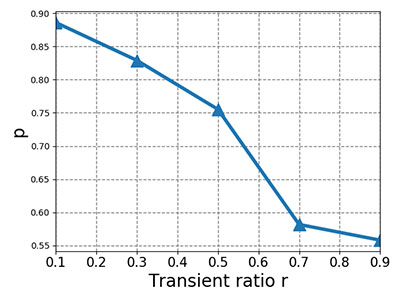}
\caption{}
\label{fig:pr}
\end{subfigure}
\begin{subfigure}{0.20\textwidth}
\centering
\includegraphics[width=\linewidth]{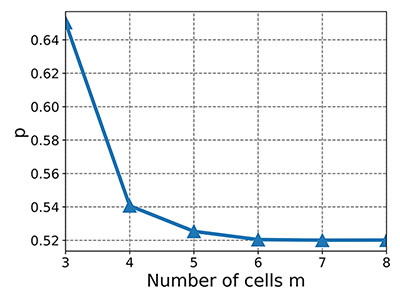}
\caption{}
\label{fig:pm}
\end{subfigure}
\caption{(a) Average $p$ value over different transient ratio $r$ (b) Average $p$ value over different number of cells $m$.} \label{fig:prpm}
\end{figure}

% p and r
Figure \ref{fig:pr} illustrates how the value of $p$ changes over different transient ratio $r$. As can be seen from the figure, when the value of $r$ increases, which means there are more transient states, the value of $p$ decrease significantly, reducing the total number of performed operations. This observation proves the functionality of portion gate that adjusts the model depth depending on sequences' latent structure.
% p and m
Figure \ref{fig:pm} shows how the value of $p$ responds to the increase in the number of cells $m$. When $m$ increase from $3$ to $8$, the average $p$ value gradually decrease, helping DA-LSTM to maintain a constant level of depth that is compatible with inputs' latent structure.

Therefore, it can be concluded that DA-LSTM architecture is capable of learning inputs' latent structure and adjusting itself by the portion gate to maintain performance.

\section{Conclusion}
In this paper, we present a novel DA-LSTM architecture to process sequential data with non-uniform information distribution. With a two-layer hierarchical structure, the capability of the model is enhanced by increasing the model depth. Additionally, our model learns the latent structure of sequential inputs and saves computation cost by dynamically adjusting the number of performed operations. Experiments conducted on real-world dataset prove that DA-LSTM architecture can preserve or even improve baseline performance sometimes, and reduce the amount of computation significantly.

\bibliographystyle{aaai}
\bibliography{refs}
\end{document}